\documentclass{article}

\usepackage[preprint]{nips_2018}

\usepackage[utf8]{inputenc} 
\usepackage[T1]{fontenc}    
\usepackage[colorlinks=true,citecolor=blue]{hyperref}       
\usepackage{url}            
\usepackage{booktabs}       
\usepackage{amsmath,amsfonts}       
\usepackage{nicefrac}       
\usepackage{microtype}      
\usepackage{graphicx}
\usepackage{color}
\usepackage{algorithm}
\usepackage{algorithmic}

\newcommand{\tr}[1]{\textrm{tr} \left( #1 \right)}
\DeclareMathOperator*{\argmin}{argmin}
\DeclareMathOperator*{\argmax}{argmax}

\title{Unsupervised Alignment of Embeddings with Wasserstein Procrustes}

\author{
  Edouard Grave \\
  Facebook AI Research \\ New York City \\
  \texttt{egrave@fb.com} \\
  \And
  Armand Joulin \\
  Facebook AI Research \\ Paris \\
  \texttt{ajoulin@fb.com} \\
  \And
  Quentin Berthet \\
  Statistical Laboratory \\ University of Cambridge \\
  \texttt{q.berthet@statslab.cam.ac.uk} \\
}

\begin{document}

\maketitle

\begin{abstract}
We consider the task of aligning two sets of points in high dimension, which has many applications in natural language processing and computer vision.
As an example, it was recently shown that it is possible to infer a bilingual lexicon, without supervised data, by aligning word embeddings trained on monolingual data.
These recent advances are based on adversarial training to learn the mapping between the two embeddings.
In this paper, we propose to use an alternative formulation, based on the joint estimation of an orthogonal matrix and a permutation matrix.
While this problem is not convex, we propose to initialize our optimization algorithm by using a convex relaxation, traditionally considered for the graph isomorphism problem.
We propose a stochastic algorithm to minimize our cost function on large scale problems.
Finally, we evaluate our method on the problem of unsupervised word translation, by aligning word embeddings trained on monolingual data.
On this task, our method obtains state of the art results, while requiring less computational resources than competing approaches.
\end{abstract}


\section{Introduction}
\label{sec:intro}

Aligning two clouds of embeddings, or high dimensional real vectors, is a
fundamental problem in machine learning with applications in natural language
processing such as unsupervised
word and sentence translation~\citep{rapp1995identifying,fung1995compiling} 
or in computer vision such as point set registration~\citep{cootes1995active} and
structure-from-motion~\citep{tomasi1992shape}.  Most of the successes of these
approaches were made in domains where either the dimension of the vectors were
small or some geometrical constraints on the point clouds were
known~\citep{fischler1987random, leordeanu2005spectral,liu2008sift}.

When dealing with unstructured sets of high dimensional vectors, it is quite common to use a few anchor points 
to learn the matching~\citep{mikolov2013exploiting, xing2015normalized}.
The supervision for this problem can be limited or noisy, for example using exact string matches in the context of word vectors alignment~\citep{artetxe2017learning}.
Recently, several unsupervised approaches have obtained compelling results, by framing this problem as some form of distance minimization between distributions,
using either the Wasserstein distance or adversarial training~\citep{cao2016distribution,zhang2017adversarial,conneau2017word}.
The methods typically require a relatively sophisticated framework that leads to a hard, and sometimes unstable, optimization problem.
Moreover, both weakly supervised and unsupervised methods greatly benefit from a refinement procedure~\citep{artetxe2017learning,conneau2017word},
often based on some variants of iterative closest points~\citep[ICP,][]{besl1992method}.
It is thus not surprising that a more direct approach, only relying on ICP, was able to achieve similar performance~\citep{hoshen2018iterative}.
However, this method is still very sensitive to the initialization, and requires a large number of random restarts, based on randomized principal component analysis to converge.

\begin{figure}
\begin{minipage}{0.65\linewidth}
\begin{center}
\includegraphics[scale=0.32]{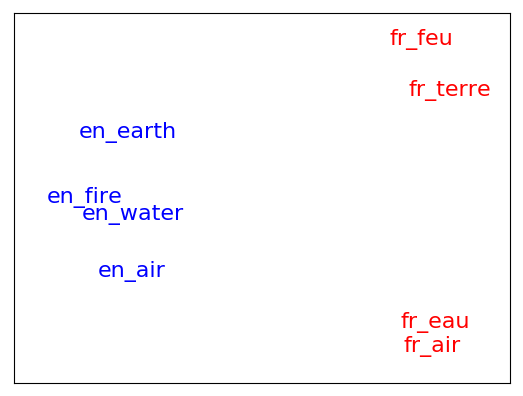}
\includegraphics[scale=0.32]{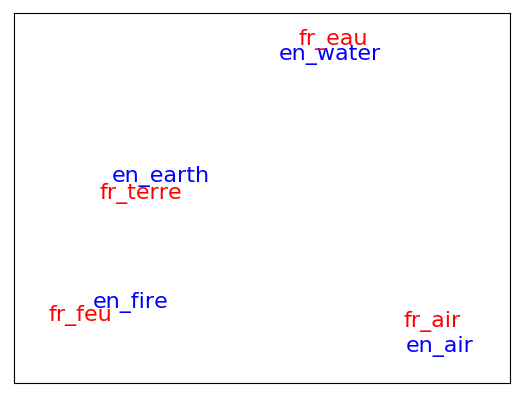}
\end{center}
\end{minipage}
\begin{minipage}{0.33\linewidth}
\caption{Illustration of the unsupervised alignment problem for word vectors.
The goal is to jointly estimate the transformation to map the vectors, as well as the correspondence between the words.
Left: PCA on the non-aligned word embeddings. Right: PCA on the word embeddings aligned with our method.
}
\end{minipage}
\end{figure}

In this work, we propose a simple approach based on jointly learning the alignment and the linear transformation between the two point clouds.
Our objective function is similar to the one of \citet{zhang2017earth} and \citet{hoshen2018iterative} but our algorithm largely differs.
In particular, our algorithm shares similarities with the work of~\citet{bojanowski2017unsupervised} where a non-linear transformation and 
an alignment between two point clouds are jointly learned.
While their goal is completely different, we take a similar optimization scheme to efficiently scale our approach to very large sets of points.
Our formulation is not convex and we propose a convex relaxation for the initialization based on a standard relaxation of the quadratic alignment
problem for graph matching~\citep{gold1996graduated}. 
Overall our approach makes little assumption about the clouds or their distributions, is flexible and converges in a few minutes.

An interesting by-product of our formulation is that it draws some similarities between
graph matching and aligning embeddings that could be used in either way. 
We validate this observation on several toy examples designed to give some insights on the 
strengths and weaknesses of our approach.
We also show that our approach is competitive with the state-of-the-art among unsupervised approaches
on word translation while running in a few minutes.
More precisely, we make the following contributions:
starting from a simple formulation, based on the joint estimation of an orthogonal matrix and a permutation matrix,
we introduce a stochastic algorithm to minimize our cost function, which can scale to large datasets ;
we show that this algorithm can be initialized using the solution of a convex problem, leading to better convergence ;
we evaluate our method on various toy experiments, as well as the task of bilingual lexicon induction.


\section{Approach}
\label{sec:approach}
In this section, we first describe Procrustes, which is a standard approach to align points when correspondences are known,
and then the Wasserstein distance which is used to measure the distance between sets of points. We then present our method
which is derived from these two techniques.

\paragraph{Procrustes.}
Procrustes analysis learns a linear transformation between two sets of matched points
$\mathbf{X} \in \mathbb{R}^{n \times d}$ and $\mathbf{Y} \in \mathbb{R}^{n \times d}$.
If the correspondences between the two sets are known (i.e., which point of $\mathbf{X}$ corresponds to which point of $\mathbf{Y}$),
then the linear transformation can be recovered by solving the least square problem:
$$
\min_{\mathbf{W} \in \mathbb{R}^{d \times d}} \| \mathbf{XW} - \mathbf{Y} \|_2^2.
$$
This technique has been successfully applied in many different fields, from
analyzing sets of 2D shapes~\citep{goodall1991procrustes} to learning a linear mapping between
word vectors in two different languages with the help of a bilingual lexicon~\citep{mikolov2013exploiting}.
Constraints on the mapping $\mathbf{W}$ can be further imposed to suit the geometry of the problem.
For example, \citet{xing2015normalized} have shown empirically that orthogonal
transformations are well suited to the mapping of word vectors.
The corresponding orthogonal Procrustes
corresponds to the following optimization problem:
\begin{eqnarray}\label{eq:orthopro}
\min_{\mathbf{Q} \in \mathcal{O}_d} \| \mathbf{XQ} - \mathbf{Y} \|_2^2,
\end{eqnarray}
where $\mathcal{O}_d$ is the set of orthogonal matrices.
This orthogonality constraint is particularly interesting since it ensures
that the distances between points are unchanged by the transformation.
As shown by~\citet{schonemann1966generalized}, the orthogonal Procrustes problem
has a closed form solution equal to $\mathbf{Q}^* = \mathbf{UV}^{\top}$, where
$\mathbf{USV}^{\top}$ is the singular value decomposition of $\mathbf{X}^{\top}\mathbf{Y}$.

\paragraph{Wasserstein distance.}
On the other hand, if we assume that the transformation between the two sets of points is known, finding the correspondences between
these sets can be formulated as the optimization problem:
$$
\min_{\mathbf{P} \in \mathcal{P}_n} \| \mathbf{X} - \mathbf{PY} \|_2^2,
$$
where $\mathcal{P}_n$ is the set of permutation matrices, i.e. the set of binary matrices
that enforces a $1$-to-$1$ mapping:
$
\mathcal{P}_n = \left\{\mathbf{P} \in \{0,1\}^{n \times n}, \hspace{1em} \mathbf{P1}_n = \mathbf{1}_n, \hspace{1em} \mathbf{P}^{\top}\mathbf{1}_n = \mathbf{1}_n \right\}.
$
Enforcing a $1$-to-$1$ mapping is not always realistic but it has the advantage to be
related to a set of orthogonal matrices.
Indeed, this makes the previous optimization problem equivalent to the following linear program:
$$
\max_{\mathbf{P} \in \mathcal{P}_n} \tr{ \mathbf{X}^{\top} \mathbf{PY}} = \max_{\mathbf{P} \in \mathcal{P}_n} \tr{ \mathbf{PYX}^{\top}}.
$$
This is known as the optimal assignment problem or
maximum weight matching problem~\citep{kuhn1955hungarian}.
It can be solved using the Hungarian algorithm, which has a complexity of $O(n^3)$.
For large number $n$ of points, the Hungarian algorithm is impractical
and an approximation is required.
For example, in the context of word translation,~\citet{zhang2017earth}
uses the fact that this problem
is equivalent to minimizing the squared Wasserstein
distance between the two sets of points $\mathbf{X}$ and $\mathbf{Y}$:
$$
W_2^2(\mathbf{X}, \mathbf{Y}) = \min_{\mathbf{P} \in \mathcal{P}_n} \sum_{i,j} P_{ij} \| \mathbf{x}_i - \mathbf{y}_j \|_2^2
\label{eq:wass}
$$
to use an approximate Earth Mover Distance solver based on the Sinkhorn algorithm~\citep{cuturi2013sinkhorn}.

\paragraph{Procrustes in Wasserstein distance.}

In our case, we do not know the correspondence between the two sets, nor the linear transformation.
Formally, our goal is thus to learn an orthogonal matrix $\mathbf{Q} \in \mathcal{O}_d$,
such that the set of point $\mathbf{X}$ is close to the set of point $\mathbf{Y}$ and $1$-to-$1$ correspondences can be inferred.
We use the Wasserstein distance defined in Eq~(\ref{eq:wass}), as the measure of distance between our two sets of points
and we combine it with the orthogonal Procrustes  defined in Eq.~(\ref{eq:orthopro}), leading to the problem of Procrustes in Wasserstein distance:
\begin{eqnarray}\label{eq:emb_align}
\min_{\mathbf{Q} \in \mathcal{O}_d} W_2^2\left(\mathbf{XQ}, \mathbf{Y} \right) =
\min_{\mathbf{Q} \in \mathcal{O}_d} \min_{\mathbf{P} \in \mathcal{P}_n} \| \mathbf{XQ} - \mathbf{PY} \|_2^2.
\end{eqnarray}
The problem is not jointly convex in $\mathbf{Q}$ and $\mathbf{P}$ but, as shown in the previous sections, there are exact solutions
for each optimization problem if the other variable is fixed.
Naively alternating the minimization in each variable does not scale and
empirical results show that it quickly converges to bad local minima even on small problems~\citep{zhang2017earth}.

\paragraph{Stochastic optimization.}

In this section, we describe a scalable optimization scheme for the problem defined in Eq.~(\ref{eq:emb_align}).  The objective of the problem in Eq.~(\ref{eq:emb_align}) can be intepreted as the Wasserstein distance $W_2^2(p^{(n)}_{\mathbf{XQ}},p^{(n)}_\mathbf{Y})$, between $p^{(n)}_{\mathbf{XQ}}$ and $p^{(n)}_\mathbf{Y}$, two empirical distributions of size $n$ of the vectors of $\mathbf{XQ}$ and $\mathbf{Y}$. One possible approach would be to alternate full minimization of $\|\mathbf{XQ-PY}\|_2^2$ in $\mathbf{P} \in \mathcal{P}_n$ and a gradient-based update in $\mathbf{Q}$. One difficulty with this method, and our formulation, is that the dimension of the permutation matrix $\mathbf{P}$ scales quadratically with the number of points $n$. Finding the optimal matching for a given orthogonal matrix has a complexity of $\mathcal{O}(n^3)$, or of $\mathcal{O}(n^2)$ up to logarithmic terms for an approximate matching via Sinkhorn \citep{cuturi2013sinkhorn,AltWeeRig17}. As a consequence, we use instead at each step $t$ a new batch of two subsamples of size $b \le n$, denoted by $\mathbf{X}_t$ and $\mathbf{Y}_t$. We compute at each step the optimal matching $\mathbf{P_t} \in \mathcal{P}_b$ and the value $W_2^2(p^{(b)}_{\mathbf{X_t Q}},p^{(b)}_{\mathbf{Y}_t})$. We then perform a gradient-guided step in $\mathbf{Q}$ for $\|\mathbf{X}_t \mathbf{Q}-\mathbf{P}_t\mathbf{Y}\|_2^2$. This surrogate objective, at each step, can be seen as a {\em subsampled} version of size $b$ of our objective. 

As in \citet{genevay2018learning}, we use samples from distributions for stochastic minimization of an objective involving a Wasserstein distance. Similarly, our algorithm can be intepreted as the stochastic optimization of a population version objective $W_2^2(p_{xQ},p_y)$, seeing $\mathbf{X,Y}$ as i.i.d. samples of size $n$ from latent unknown distributions $p_{x}$ and $p_y$, and $\mathbf{X}_t, \mathbf{Y}_t$ as samples of size $b$. At a high level, this approach can be motivated by the convergence of $W_2(\mu_m,\nu_m)$ to $W_2(\mu,\nu)$, where $\mu_m$ and $\nu_m$ are the empirical distributions of i.i.d. samples of size $m$ from latent distributions $\mu$ and $\nu$, via the analysis of $W_2(\mu_m,\mu)$ and the triangle inequality. Results of these type, and analysis of the rates of convergence go back to \cite{dudley1969speed} for the $W_1$ distance, and have been extended to all $W_p$ distances, including $p=2$ \citep{weed2017sharp}, the case of discrete distributions \citep{TamSomMun17,SomSchMun18}, and the use of these methods in variational problems \citep{BerJacGer17}. While these results do not give formal guarantees for our approach, they provide some motivation for the success of this method.

Optimizing a general function over the manifold of the orthogonal matrices, the \emph{Stiefel manifold},
as been thoroughly studied~\citep{absil2009optimization}.
A simple solution is to take a step in a descending direction (given the full gradient or a stochastic approximation)
while pulling back the update on the manifold.
Most of the pull back operators are computationally expensive but our matrix $\mathbf{Q}$ is relatively small.
In particular, we use the projection even though it requires a singular value decomposition as shown in the previous section.
Overall, our optimization in $\mathbf{Q}$ is a stochastic gradient descent (SGD) with a projection on the Stiefel manifold.

\begin{minipage}{0.33\linewidth}
The overall optimization scheme is summarized in Algorithm 1.
At the iteration $t$, we sample a mini-batch $\mathbf{X}_t \in \mathbb{R}^{b \times d}$ and $\mathbf{Y}_t \in \mathbb{R}^{b \times d}$ of points
of size $b$ from the matrices $\mathbf{X}$ and $\mathbf{Y}$.
We then compute the optimal permutation for this batch by solving the same problem as in Eq.~(\ref{eq:wass}).
Given this permutation, we compute the gradient in $\mathbf{Q}$ to update the matrix.
While this procedure is efficient and can scale to very large sets of points, it is not convex and the quality of its
solution depends on its initialization.
In particular, the initial point $\mathbf{Q}_0$ is very important for the quality of the final solution.
In the next section, we discuss a strategy to initialize this optimization problem.
\end{minipage} \hfill
\begin{minipage}{0.65\linewidth}
\vspace{-1.2em}
\begin{algorithm}[H]
\caption{Stochastic optimization}
\begin{algorithmic}[1]
\FOR{$t = 1$ to $T$}
\STATE Draw $\mathbf{X}_t$ from $\mathbf{X}$ and $\mathbf{Y}_t$ from $\mathbf{Y}$, of size $b$
\STATE Compute the optimal matching between $\mathbf{X}_t$ and $\mathbf{Y}_t$ given the current orthogonal matrix $\mathbf{Q}_t$
$$
\mathbf{P}_t = \argmax_{\mathbf{P} \in \mathcal{P}_b} \tr{\mathbf{Y}_t \mathbf{Q}_t^{\top} \mathbf{X}_t^{\top} \mathbf{P}}.
$$
\vspace{-0.5em}

\STATE Compute the gradient $\mathbf{G}_t$ with respect to $\mathbf{Q}$:
$$ \mathbf{G}_t = -2\mathbf{X}_t^{\top} \mathbf{P}_t \mathbf{Y}_t. $$
\vspace{-1.0em}

\STATE Perform a gradient step and project on the set of orthogonal matrices:
$$
\mathbf{Q}_{t+1}~=~\Pi_{\mathcal{O}_d} \left( \mathbf{Q}_t - \alpha \mathbf{G}_t \right).
$$
\vspace{-0.5em}

For a matrix $\mathbf{M} \in \mathbb{R}^{d \times d}$, the projection is given by
$ \Pi_{\mathcal{O}_d}(\mathbf{M}) = \mathbf{UV}^{\top}, $
with $\mathbf{USV}^{\top}$ the singular value decomposition of $\mathbf{M}$.
\ENDFOR
\end{algorithmic}
\end{algorithm}
\end{minipage}

\paragraph{Convex relaxation.}
In this section, we propose a convex relaxation of the problem defined in Eq.~(\ref{eq:emb_align}).
This relaxation comes from the observation that our problem is equivalent to
$$
\max_{\mathbf{P} \in \mathcal{P}_n} \max_{\mathbf{Q} \in \mathcal{O}_d} \tr{ \mathbf{Q}^{\top} \mathbf{X}^{\top} \mathbf{P Y}}.
$$
Solving a linear program over $\mathcal{O}_d$ is equivalent to solving it as over its convex hull, i.e.,
the set of matrices with a spectral norm lower than $1$.
This value at this maximum is thus equal to the dual norm of the spectral norm, i.e., the trace norm (or nuclear norm) of $\mathbf{X}^{\top} \mathbf{PY}$.
Thus, the problem is equivalent to:
$$
\max_{\mathbf{P} \in \mathcal{P}_n} \| \mathbf{X}^{\top} \mathbf{PY} \|_*,
$$
where $\mathbf{Z} \mapsto \| \mathbf{Z} \|_*$ is the trace norm.
The trace norm requires to compute the singular values of the matrix $\mathbf{X}^{\top} \mathbf{PY}$, which is computationally expensive.
Another possible formulation is to replace the trace norm by the Frobenius norm, leading to the following:
\begin{align}
\max_{\mathbf{P} \in \mathcal{P}_n} \| \mathbf{X}^{\top} \mathbf{PY} \|_2^2 
& = \max_{\mathbf{P} \in \mathcal{P}_n} \tr{ \mathbf{K_Y P}^{\top} \mathbf{K_X P}},
\end{align}
where $\mathbf{K_X} = \mathbf{XX}^{\top}$ and $\mathbf{K_Y} = \mathbf{YY}^{\top}$.
This formulation has the advantage of leading to a quadratic assignment program, i.e.,
\begin{eqnarray}
\min_{\mathbf{P} \in \mathcal{P}_n} \| \mathbf{K_X P} - \mathbf{P K_Y} \|_2^2,
\end{eqnarray}
since the permutation matrices are orthogonal, and the Frobenius norm is invariant by multiplication with an orthogonal matrix.
This problem is a standard formulation for the graph isomorphism, or graph matching problem
and it is known to be NP-hard in general~\citep{garey2002computers}.
However, many convex relaxations have been proposed for this problem.
Of particular interest, \citet{gold1996graduated} replace the set of permutation matrices by its convex hull,
the set of doubly stochastic matrices, namely the Birkhoff polytope, leading to the following convex relaxation:
\begin{eqnarray}\label{eq:graph_match}
\min_{\mathbf{P} \in \mathcal{B}_n} \| \mathbf{K_X P} - \mathbf{P K_Y} \|_2^2,
\end{eqnarray}
where $\mathcal{B}_n = \textrm{convex-hull}(\mathcal{P}_n)$ is the Birkhoff polytope.
It would be interesting to see what would be the equivalent relaxation for the trace norm but this goes beyond the scope of this paper.

We use the Frank-Wolfe algorithm to minimize this problem~\citep{frank1956algorithm}. Once the global minimizer $\mathbf{P}^*$ has been
attained, we compute a corresponding orthogonal matrix $\mathbf{Q}_0$ by solving:
$$
\mathbf{Q}_0 = \argmin_{\mathbf{Q} \in \mathcal{O}_d} \| \mathbf{X Q} - \mathbf{P}^* \mathbf{Y} \|_2^2.
$$
This corresponds to taking the singular value $\mathbf{USV}^{\top}$ of $\mathbf{P}^* \mathbf{Y X}^{\top}$, and setting $\mathbf{Q}_0=\mathbf{UV}^{\top}$.
Note that the matrix $\mathbf{P}^*$ is not necessarily a permutation matrix, but only doubly stochastic.

\paragraph{Improving the nearest-neighbor search.}
Once the source embeddings $\mathbf{X}$ are mapped to the target space, they are not
perfectly aligned with target embeddings $\mathbf{Y}$ and a retrieval procedure is required.
In high dimensional vector space, some points, called \emph{hubs}, tend to be close to disproportionally many vectors and a direct nearest-neighbor (NN) search
favors these hubs~\citep{aucouturier2008scale,radovanovic2010hubs}.
This problem has been observed in many retrieval situations~\citep{doddington1998sheep,pachet2004improving,dinu2014improving}.

Several strategies have been proposed to diminish the effects of hubs, most are based
on some definition of a local distance~\citep{jegou2010accurate,conneau2017word}
or by reversing the direction of the retrieval~\citep{dinu2014improving,smith2017offline}.
In this paper, we consider the Inverted Softmax (ISF) proposed by~\citet{smith2017offline} and the Cross-Domain Similarity Local Scaling (CSLS)
of~\citet{conneau2017word}.
The ISF is defined for normalized vectors as:
$$\text{ISF}(\mathbf{y}, \mathbf{z}) = \frac{\exp(\beta \mathbf{y}^\top \mathbf{z})}{\sum_{\mathbf{y}'\in \mathbf{Y}}\exp(\beta \mathbf{y}^{'\top}\mathbf{z})},$$
where $\beta>0$ is a temperature parameter.
CSLS is a similarity measure between the vectors $\mathbf{y}$ and $\mathbf{z}$ from 2 different sets $\mathbf{Y}$ and $\mathbf{Z}$, defined as
$\text{CSLS}(\mathbf{y}, \mathbf{z}) = 2 \textrm{cos}(\mathbf{y}, \mathbf{z}) - R_\mathbf{Z}(\mathbf{y}) - R_\mathbf{Y}(\mathbf{z}),$
where $\textrm{cos}(\mathbf{y}, \mathbf{z}) = \frac{\mathbf{y}^{\top}\mathbf{z}}{\|\mathbf{y}\| \|\mathbf{z}\|}$ is the cosine similarity between $\mathbf{y}$ and $\mathbf{z}$, and $R_\mathbf{Z}(\mathbf{y})=\frac{1}{K}\sum_{\mathbf{z}\in\mathcal{N}_Z(\mathbf{y})} \textrm{cos} (\mathbf{z}, \mathbf{y})$ is the average of the cosine similarity between $\mathbf{y}$ and its $K$ nearest neighbors among the vectors in $\mathbf{Z}$.
It is an extension of the contextual dissimilarity measure of~\citet{jegou2010accurate} to two sets of points.
An advantage of the CSLS measure over ISF is that its free parameter (i.e., the number of nearest neighbors) is much simpler to set than
the temperature of the ISF.


\section{Related work}
\label{sec:related_work}

\paragraph{Embedding alignment with no bilingual dictionary.}
In the case where the permutation matrix is given by a bilingual dictionary,
the problem stated in Eq.~(\ref{eq:emb_align}) has been studied in the context of word translation by~\citet{mikolov2013exploiting} 
with $\mathbf{Q}$ as an unconstrained linear transformation.
\citet{xing2015normalized} have shown that constraining $\mathbf{Q}$ to be an orthonormal matrix yields better results if used with normalized embeddings,
and \citet{artetxe2016learning} have improved further the quality of supervised alignment by using centered normalized embeddings.
\citet{hoshen2018iterative} is the closest attempt to learn unsupervised word translation with a formulation similar to Eq.~(\ref{eq:emb_align}).
They use an iterative closest point approach initialized with a randomized PCA.
Others have looked into unsupervised aligning word embedding distributions instead of an explicit alignment: 
\citet{cao2016distribution} align the moments of the two distributions which assumes a Gaussian distribution over the embeddings,
while others~\citep{zhang2017adversarial, conneau2017word} have preferred the popular Generative Adversarial Networks (GANs) framework 
of~\citet{goodfellow2014generative}.
Recently, \citet{zhang2017earth} have proposed to use the earth mover distance to refine the alignment of word embeddings obtained
from a generative adversarial network.

\paragraph{Graph matching.}
The graph matching formulation of Eq.~(\ref{eq:graph_match}) 
is a quadratic assignment problem, that is NP-Hard~\citep{garey2002computers}.
Polynomial algorithms exist for restricted classes of graphs, like planar graphs~\citep{hopcroft1974linear} or trees~\citep{aho1974design}, and for the general case, 
many heuristics have been proposed, some can be found in the review papers 
of~\citet{conte2004thirty} and~\citet{foggia2014graph}.
Closer to our approach is the convex relaxation to a linear program proposed in
\citet{almohamad1993linear} and even more, the convex relaxation proposed by \citet{gold1996graduated} of
 the set of permutation matrices to its convex hull, i.e., the set of
doubly-stochastic matrices.
Under certain conditions on the graph structures, 
the solution of this relaxation is the same as the exact isomorphism~\citep{aflalo2014graph}, but
in most cases, there are no guarantees to converge to 
a corner of the convex hull, and can even provably converge to poor solutions
for some graph families~\citep{lyzinski2016graph}.
However, \citet{lyzinski2016graph} have suggested that they can be used as an
initialization for an inexact or non-convex algorithm like the concave
minimization problem of~\citet{zaslavskiy2009path} or the problem
formulated in Eq.~(\ref{eq:emb_align}).

\paragraph{Wasserstein distance.}
As in~\citet{bojanowski2017unsupervised}, our formulation in
Eq.~(\ref{eq:emb_align}) minimizes the Wasserstein distance (also known as Earth Mover's distance) to estimate 
the minimal transformation between two distributions of points~\citep{rubner1998metric}.
We refer the reader to~\citet{peyre2017} for an exhaustive survey on the subject. 
\citet{kusner2015word} use an EMD between word embeddings to compute the distance between documents,
and later~\citet{rolet2016fast} use a smooth Wasserstein distance to align word embeddings distribution and 
discover cross-language topics. \citet{FlaCutCou16} optimize a different Wasserstein objective over the Stiefel manifold, for the task of discriminant analysis. The Wasserstein distance has also been used in the field of statistics to give meaningful notions of means for misaligned signals \citep{PanZem16,ZemPan17}.
Finally, our approach bears some similarities with the
Gromov-Wasserstein that has been successfully used to align shapes
in~\citet{memoli2007use,solomon2016entropic}.


\section{Experiments}
\label{sec:results}

In this section, we evaluate our method in two settings.
First, we perform toy experiments to gain some insight regarding our approach.
We then compare it to various state of the art algorithms on the task of unsupervised word translation.

\begin{table}[t]
\begin{minipage}{0.6\linewidth}
\centering
\begin{tabular}{lcccc}
\toprule
         & Seed  & Data  & Window & Algorithm \\
\midrule
Distance & 6.090 & 7.872 & 11.151 & 16.008    \\
\midrule
Relaxation     & 0.39 & 0.23 & 0.00  & 0.00 \\
Relaxation$^*$ & 0.99 & 0.97 & 0.22  & 0.00 \\
Random init    & 0.70 & 0.62 & 0.39  & 0.59 \\
Convex init    & 1.00 & 1.00 & 0.987 & 0.985 \\
\midrule
\end{tabular}
\end{minipage}
\begin{minipage}{0.38\linewidth}
\caption{We report the accuracy of different methods for words with rank in the range 5,000-10,000.
Relaxation indicates is the convex formulation applied to 2k random points, while relaxation$^*$ indicates the convex formulation applied to the 2k first vectors from each set.}
\label{tab:toy}
\end{minipage}
\end{table}

\subsection{Toy experiments}

Instead of generating purely random datasets, we train various word embeddings, using the same corpus but introducing some noise in the training process.
More specifically, we train skipgram and cbow models using the \texttt{fasttext} tool~\citep{bojanowski2016enriching} on 100M English tokens from the 2007 News Crawl corpus.\footnote{http://statmt.org/wmt14/translation-task.html}
We consider the following settings to generate our toy datasets:
\begin{itemize}
\item \textbf{Seed:} we learn two models on the same data, with the same hyper-parameters.
The only difference between the two models is the seed used to initialize the parameters of the models.
Please note that during the training of skipgram models, frequent words are randomly subsampled, as well as negative examples.
There is thus two sources of randomness between the two models: the initial parameters of the model and the sampling during training.
\item \textbf{Data:} in that instance, we learn two embeddings using different split of the data.
More precisely, we use the first 100M tokens for the first model and the next 100M tokens for the second model.
Since both training splits come from the same domain, the models are closed.
\item \textbf{Window:} in that instance, we learn two skipgram models on the same data, but with different window size.
The first model uses a window of size $2$, while the second model uses a window of size $10$.
We also use different seeds, similarly to the first instance.
\item \textbf{Algorithm:} for the last setting, we consider two word embeddings from different models: skipgram and cbow.
\end{itemize}
Because all models are trained on English data, we have the ground truth matching between vectors from the two sets.
We can thus estimate an orthogonal matrix using Procrustes, and measure the ``distance'' between the two sets of points.
We report these for the different settings in Table~\ref{tab:toy}.
For these experiments, we compare the different approaches on the first 10,000 points of our models.

First, we compare our stochastic algorithm with random initialization with the same algorithm initialized with the solution of the convex relaxation.
We observe that while the random initialization sometimes converges for the easy problems, its rate of success rapidly decreases.
On the other hand, when initialized with the solution of the convex formulation, our stochastic algorithm always converges to a good solution.
Interestingly, even when the solution of the relaxed problem is not a good solution, it still provides a good initialization point for the stochastic gradient method.

\begin{table}[t]
\centering
\begin{tabular}{l cc cc cc cc}
\toprule
 & \textsc{en-es} & \textsc{es-en} & \textsc{en-fr} & \textsc{fr-en} & \textsc{en-de} & \textsc{de-en} & \textsc{en-ru} & \textsc{ru-en} \\
\midrule
Procrustes         & 82.7 & \underline{84.2} & \underline{82.7} & \underline{83.4} & 74.8 & 73.2 & \underline{51.3} & \underline{63.7} \\ 
\midrule
Adversarial$^*$    & 81.7 & 83.3 & 82.3 & 82.1 & 74.0 & 72.2 & 44.0 & 59.1 \\
ICP$^*$            & 82.1 & \textbf{84.1} & 82.3 & \textbf{82.9} & 74.7 & 73.0 & \textbf{47.5} & \textbf{61.8} \\
Ours               & \underline{\textbf{82.8}} & \textbf{84.1} & \textbf{82.6} & \textbf{82.9} & \underline{\textbf{75.4}} & \underline{\textbf{73.3}} & 43.7 & 59.1 \\
\bottomrule
\end{tabular}
\caption{Performance of different methods using refinement.
In underline the best performance, in bold, the best among unsupervised methods.
$^*$ indicates unnormalized vectors.}\label{tab:word_trans}
\end{table}

\subsection{Unsupervised word translation}
In our second set of experiments, we evaluate our method on the task of unsupervised word translation.
Given word embeddings trained on two monolingual corpora, the goal is to infer a bilingual dictionary by aligning the corresponding word vectors.
We use the same exact setting as~\citet{conneau2017word}. In particular, we use the same evaluation datasets and code, as well as the same word vectors.

\paragraph{Baselines.}
We compare our method with Procrustes, as well as two unsupervised approaches:
the adversarial training (adversarial) of~\citet{conneau2017word}
and the Iterative Closest Point approach (ICP) of~\citet{hoshen2018iterative}.
All their numbers are taken from their papers.

\paragraph{Implementation details.}
For CSLS, we use $10$ nearest neighbors and for ISF, we use a temperature of $25$.
Following~\citet{xing2015normalized} and \citet{artetxe2016learning}, we normalize and center the embeddings for both Procrustes and our approach.
The refinement approach is taken from~\citet{conneau2017word} and consists in alternating, for $5$ epochs,
between building a dictionary of mutual nearest neighbors (using CSLS) and running Procrustes on this dictionary.
This procedure is thus ICP with the CSLS criterion.
Our convex relaxation is too demanding to run on all the vectors, and thus, we use the 2.5K most frequent words for our initialization.
As noted in section~\ref{sec:approach}, the choice of the batch size is important: a larger batch size will lead to a better
approximation of our cost function, but is also more computationally intensive.
Thus, we start with a small batch size of $500$, and double the batch size after a third and two third of the optimization.
Finally, we use the Sinkhorn solver of \citet{cuturi2013sinkhorn} to compute approximate solutions of optimal transport problems.

\paragraph{Main results.}
In order to quantitatively assess the quality of each approach, we consider the problem of bilingual lexicon induction.
This can be formulated as a retrieval problem, and following standard practice, we report the precision at one.
As observed by \citet{artetxe2017learning} and \citet{conneau2017word}, refining the alignment significantly improves the performance of unsupervised approaches.
We thus report the performance of different methods after refinement in Table~\ref{tab:word_trans}.
It is interesting to see that the refinement also improves supervised approaches.
Several reasons can explain this: first the lexicons for supervision
are not clean, and the $\ell_2$-loss is sensible to outliers. Second the lexicons
are small (around $10$K) and the mapping is not regularized, besides the orthogonality constraint.

\begin{table*}
\centering
\begin{tabular}{l cc cc cc cc}
\toprule
Method &\textsc{en-es}&\textsc{es-en}&\textsc{en-fr}&\textsc{fr-en}&\textsc{en-de}&\textsc{de-en}&\textsc{en-ru}&\textsc{ru-en}\\
\midrule
Proc. - NN    & 78.3 & 80.9 & 77.0 & 80.0 & 70.8 & 71.8 & 49.5 & 64.6 \\
Proc. - CSLS  & 81.1 & \underline{84.8} & \underline{81.6} & \underline{84.0} & \underline{74.7} & \underline{74.4} & \underline{52.0} & \underline{67.8} \\
Proc. - ISF   & \underline{81.4} & 83.4 & 81.2 & 82.9 & 71.8 & 72.4 & 50.4 & \underline{67.8} \\
\midrule
Adv. - NN     & 69.8 & 71.3 & 70.4 & 61.9 & 63.1 & 59.6 & 29.1 & 41.5 \\
Adv. - CSLS   & 75.7 & 79.7 & 77.8 & 71.2 & \textbf{70.1} & \textbf{66.4} & 37.2 & 48.1 \\
\midrule
Ours - NN             & 77.2 & 75.6 & 75.0 & 72.1 & 66.0 & 62.9 & 32.6 & 48.6 \\
Ours - CSLS           & 79.8 & \textbf{81.8} & \textbf{79.8} & \textbf{78.0} & 69.4 & \textbf{66.4} & \textbf{37.5} & \textbf{50.3} \\
Ours - ISF            & \textbf{80.2} & 80.3 & 79.6 & 77.2 & 66.9 & 64.2 & 36.9 & \textbf{50.3} \\
\bottomrule
\end{tabular}
  \caption{Comparison between supervised and unsupervised word translation with no refinement. In underline the best performance, in bold, the best among unsupervised approaches.
  We normalize and center the word embeddings for Proc. and our approach, while Adv. uses unnormalized vectors.
}\label{tab:raw}
\end{table*}

Overall our performance is on par with ICP and significantly better than adversarial training.
The fact that ICP and our approach achieve comparable results is not surprising as we are considering a similar loss function.
However, ICP requires a lot of random initializations to converge to a good solution,
while initializing our approach with the result of the convex relaxation guarantees the
same performance with a single run.

\paragraph{Results without refinement.}
The refinement step used in Table~\ref{tab:word_trans} is an ICP with a CSLS criterion.
This procedure requires a good initialization to converge to a satisfying solution.
\citet{hoshen2018iterative} uses hundreds of initializations to find a good starting point, while the adversarial training often gives a decent initialization with a single run.
We compare the quality of this intialization with the solution of our relaxed method in Table~\ref{tab:raw}.
For both NN and CSLS, our approach performs better than the GAN based method of~\citet{conneau2017word} on $6$ out of $8$ pairs of languages while being simpler.
However, we are still significantly worse than the supervised approach, by often more than $5\%$.
It is interesting to notice that for both Procrustes and our approach, ISF and CSLS obtain similar performance.
Empirically, we found that ISF is more sensitive to its free parameter.
In particular, the temperature in ISF should vary with the number of points,
and the fact that a single value works across all these datasets is potentially
an indirect consequence of the evaluation, that is to restrict the
size of the sets to $200K$ words.
We thus propose to use CSLS in most of our experiments.

\begin{table}[t]
\begin{minipage}{0.65\linewidth}
\centering
\begin{tabular}{lccccc}
\toprule
                   & 100  & 200  & 400  & 800  & 1600 \\
\midrule
Time             & 1m47s & 2m07s & 2m54s & 5m34s & 22m13s \\
\midrule
\textsc{en-es}   & 68.5 & 73.8 & 74.9 & 75.0 & 76.3 \\
\textsc{en-fr}   & 67.4 & 71.9 & 74.5 & 75.6 & 75.7 \\
\textsc{en-de}   & 59.1 & 63.0 & 64.4 & 65.8 & 66.4 \\
\textsc{en-ru}   & 23.7 & 27.9 & 29.9 & 32.3 & 33.2 \\
\midrule
\end{tabular}
\end{minipage}
\begin{minipage}{0.33\linewidth}
\caption{Influence of the batch size: we report the precision at 1 after 4,000 iterations as a function of the batch size.
We use the nearest neighbor (NN) approach to retrieve the translation of a given query.}
\label{tab:bsz}
\end{minipage}
\end{table}

\paragraph{Impact of the batch size on the performance.}
The size of the batch size $b$ plays an important role on our surrogate loss as it trades off speed for distance to the original formulation.
Table~\ref{tab:bsz} shows its influence on the performance as well as running time. As expected, our model converges faster for small $b$
since we run the Sinkhorn algorithm on $b\times b$ matrices, thus leading to a complexity of $\mathcal{O}(b^2)$ up to logarithmic terms \citep{AltWeeRig17}.
The larger $b$ is, the better the approximation of the squared Wasserstein distance is, and the closer we are to the real loss function.
Despite saying nothing about the quality of the local minima reached by the surrogate,
it is interesting to see that in practice this converts into better performance as well.

\section{Conclusion}

This paper presents a general approach to align embeddings in high dimensional space.
While the overall problem is non-convex and computationally expensive, we present an efficient
stochastic algorithm to solve the problem.
We also develop a convex relaxation that can be used to initialize our approach.
We validate our method on a few toy examples and a real application, namely unsupervised
word translation, where
we achieve performances that are on par with the state-of-the-art.
A few questions remain, in particular the link between graph matching and point cloud alignments
can be further investigated. It should be possible to identify the type of problems where our
approach is guaranteed to work and where it will provably fail.
Finally, we think that it is possible to improve the relaxation procedure.

\bibliography{example_paper}
\bibliographystyle{apalike}

\end{document}